\documentclass[letterpaper]{article} 
\usepackage{aaai2026}  
\usepackage{times}  
\usepackage{helvet}  
\usepackage{courier}  
\usepackage[hyphens]{url}  
\usepackage{graphicx} 
\usepackage{natbib}  
\usepackage{caption} 
\usepackage{algorithm}       
\usepackage{algpseudocode}

\usepackage{tikz}
\usetikzlibrary{positioning, arrows.meta}
\usepackage{booktabs}
\usepackage{xcolor}
\usepackage{soul}
\usepackage{placeins}  
\usepackage{amsmath}
\usepackage{tabularx}

\usepackage{newfloat}
\usepackage{listings}
\DeclareCaptionStyle{ruled}{labelfont=normalfont,labelsep=colon,strut=off} 
\floatstyle{ruled}
\newfloat{listing}{tb}{lst}{}
\floatname{listing}{Listing}
\lstdefinelanguage{json}{
  basicstyle=\ttfamily\footnotesize,
  showstringspaces=false,
  breaklines=true,
  morestring=[b]",
  morekeywords={true,false,null}
}

\lstdefinestyle{jsonstyle}{
  language=json,
  numbers=left,
  numbersep=6pt
}

\usepackage{eso-pic}  

\title{Improving Procedural Skill Explanations via Constrained Generation: A Symbolic-LLM Hybrid Architecture}
\author {
    Rahul Dass,
    Thomas Bowlin,
    Zebing Li,
    Xiao Jin and 
    Ashok Goel
}
\affiliations {
    Georgia Institute of Technology \\
    \{rdass7,tbowlin3,zebing.li,xjin96,ag25\}@gatech.edu
}

\usepackage{bibentry}

\begin{document}

\maketitle

\begin{abstract}
In procedural skill learning, instructional explanations must convey not just steps, but the causal, goal-directed, and compositional logic behind them.
Large language models (LLMs) often produce fluent yet shallow responses that miss this structure.
We present Ivy, an AI coaching system that delivers structured, multi-step explanations by combining symbolic Task-Method-Knowledge (TMK) models with a generative interpretation layer-an LLM that constructs explanations while being constrained by TMK structure.
TMK encodes causal transitions, goal hierarchies, and problem decompositions, and guides the LLM within explicit structural bounds.
We evaluate Ivy against responses against GPT and retrieval-augmented GPT baselines using expert and independent annotations across three inferential dimensions.
Results show that symbolic constraints consistently improve the structural quality of explanations for ``how'' and ``why'' questions.
This study demonstrates a scalable AI for education approach that strengthens the pedagogical value of AI-generated explanations in intelligent coaching systems.
\end{abstract}


\section{Introduction}
In Spring 2025, Ivy \cite{dass2025ivy} is an AI coaching system that was deployed in an online graduate-level AI course at a large U.S. university [Redacted for double-blind review], where it answered students' questions about procedural skills.
Across multiple pilot studies, Ivy's responses were perceived fluent and often helpful, yet we observed a recurring gap: explanations can lack the structured logic instructors expect for teaching procedural skills-how steps depend on state (\textit{causality}), why steps serve an overarching goal (\textit{teleology}), and how problems decompose into subgoals (\textit{decomposition}) \cite{lum2025designing}.

This gap matters educationally.
Learners increasingly rely on AI-generated explanations when practicing procedural tasks like writing code, debugging, or constructing arguments.
But skill transfer, the benchmark of meaningful learning, depends not only on surface fluency but on understanding \textit{why} and \textit{how} procedures work \cite{bransford2000people}.
Transfer requires causal insight into the logic of tasks-a hallmark of expert reasoning that unconstrained LLMs may fail to replicate \cite{bransford2000people}.
Thus, improving the inferential structure of explanations is central to pedagogical value, not a peripheral nicety.

Currently, Ivy uses Task-Method-Knowledge (TMK) models \cite{goel2017gaia}, we refer to as ``TMK-Basic'': they capture goals and associated mechanisms but do not explicitly encode causal transitions, goal hierarchies, or formal goal decomposition.
In this work, we introduce ``TMK-Structured'', which makes these three inferencing structures explicit: causal transitions (modeled through finite state machines), teleological linkages (via goal hierarchies), and hierarchical decomposition (through goal abstraction)-each capturing a different aspect of how a goal unfolds, why it matters, and how it can be broken down.

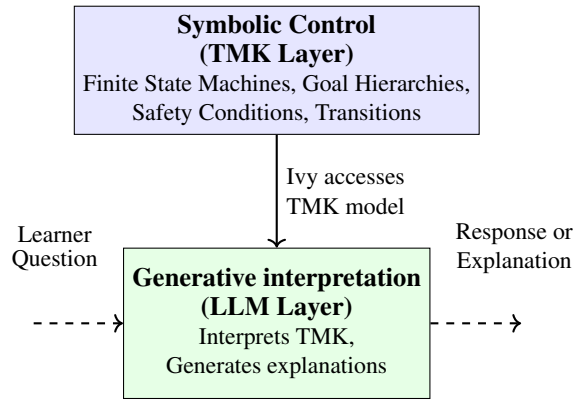
\begin{figure}[h]
\centering
\begin{tikzpicture}[node distance=1.5cm, every node/.style={align=center}]
  \node (TMK) [draw, rectangle, minimum width=1.2cm, minimum height=1cm, fill=blue!10] {%
    \textbf{Symbolic Control} \\ 
    \textbf{(TMK Layer)}\\
    \small Finite State Machines, Goal Hierarchies,\\
    \small Safety Conditions, Transitions
  };

  \node (llm) [below=of TMK, draw, rectangle, minimum width=1.2cm, minimum height=2cm, fill=green!10] {%
    \textbf{Generative interpretation} \\ 
    \textbf{(LLM Layer)}\\
    \small Interprets TMK,\\
    \small Generates explanations
  };

  \draw[->, thick] (TMK) -- node[right, align=left] {\footnotesize \shortstack{Ivy accesses \\ TMK model}} (llm);
  \draw[<-, thick, dashed] (llm) -- ++(-3.25, 0) node[left, xshift=1cm, yshift=1cm] {\footnotesize \shortstack{Learner\\Question}};
  \draw[->, thick, dashed] (llm) -- ++(3.25, 0) node[right, xshift=-1cm, yshift=1cm] {\footnotesize \shortstack{Response or \\Explanation}};
\end{tikzpicture}
\caption{Concept diagram of Ivy's two-layered inferencing: TMK encodes structured knowledge; LLMs perform runtime inference within these constraints.}
\label{fig:causal_teleo_reasoning}
\end{figure}

We present a constrained generation architecture that separates symbolic control (TMK layer) from generative interpretation and synthesis (LLM layer), see Figure \ref{fig:causal_teleo_reasoning}.

We evaluate Ivy+TMK-Structured against three baselines, including Ivy+TMK-Basic using expert and independent ratings of correctness and three inferential dimensions.
Our primary hypothesis is that TMK-Structured will increase the prevalence and explicitness of these inferential structures in generated explanations relative to the baselines.

Together, the architecture and evaluation advance a practical question for AI in education: can symbolic procedural constraints make LLM explanations not only more reliable, but also more pedagogically aligned with the way instructors teach procedural skills?
We observe promising gains, most notably in problem decomposition-the dimension most directly scaffolded by the TMK-Structured design.


\section{Related Work}
\subsection{Structured Reasoning and Transfer}
Long-standing work in the learning sciences shows that successful skill transfer requires more than surface fluency; learners must understand underlying principles and know when/how/why to apply them  \cite{bransford2000people}.
Causal understanding in particular distinguishes expert reasoning and supports generalization across tasks \cite{bransford2000people,ritscher2003understanding}, while compositional skill representations help learners adapt knowledge to new contexts \cite{lamri2023reconciling}.
Although instruction emphasizes mechanistic (cause-and-effect) explanations, teleological biases can reappear even in trained adults under cognitive load \cite{kelemen2009human}; targeted interventions can improve mechanistic reasoning \cite{gonzalezgalli2022impact}, and distinguishing valid purpose-based accounts from misconceptions remains pedagogically important \cite{kampourakis2020students}. 
Our work operationalizes these insights by explicitly scaffolding causal transitions, goal–mechanism linkages, and problem decompositions in generated explanations.

\subsection{Constrained Generation for Multi-Step Inferencing}
While retrieval augmented generation (RAG) improve relevance, it does not guarantee structured reasoning in explanations \cite{gao2023survey}. 
Recent methods add formal constraints to LLM decoding: CRANE uses domain grammars to improve symbolic QA \cite{banerjee2025crane}, and grammar-constrained decoding can outperform in-context examples on logic tasks \cite{raspanti2025grammar}. 
Retrieval-Augmented Thoughts (RAT) iterates retrieval and reasoning to elicit multi-hop chains \cite{wang2024rat}. 
Our approach complements these lines by separating symbolic control from generative synthesis: structured TMK representations define the inferential space, and the LLM synthesizes explanations within those constraints.

\subsection{Hybrid Symbolic-Neural Architectures for Reducing Hallucinations}
Surveys of knowledge-aware LLM pipelines find structured grounding reduces hallucinations and benefits smaller models in particular \cite{agrawal2024can}.
Domain systems combining knowledge graphs with LLMs improve factuality (e.g., KGT in biomedical QA) \cite{feng2025knowledge}, and structured-output RAG reduces hallucinations and supports out-of-domain generalization \cite{bechard2024reducing}.
Complementary detection methods use mechanistic signals to flag hallucinations (e.g., ReDeEP) \cite{sun2025redeep}.
Unlike approaches that treat structure as static context, we use symbolic representations as active generative constraints, yielding explanations that are both more reliable and more pedagogically structured.

\begin{table}[t]
\centering
\renewcommand{\arraystretch}{1.2}
\resizebox{1.0\linewidth}{!}{  
    \begin{tabular}{p{2.2cm} p{6.2cm}}
    \toprule
    \textbf{TMK-Structured Schema Field} & \parbox[t]{6.2cm}{\textbf{Guards And Prisoners}\\\textbf{Example}} \\
        \midrule
    name & \texttt{ReturnGuardMechanism} \\
    inputParameters & {\small\texttt{[b: boat, g: guard, c: configuration]}} \\
    outputParameters & {\small\texttt{[newConfig: configuration]}} \\
    startState & \texttt{\textbf{RG\_S0}} \\
    successState & \texttt{\textbf{RG\_S3}} \\
    failureState & \texttt{\textbf{RG\_Fail}} \\
    states &
    \texttt{\textbf{RG\_S0}: \textcolor{blue}{EmbarkGuard}} (operation) \\
    & \texttt{\textbf{RG\_S1}: \textcolor{blue}{Cross}} (operation) \\
    & \texttt{\textbf{RG\_S2}: \textcolor{blue}{DisembarkGuard}} (operation) \\
    & \texttt{\textbf{RG\_S3}: \textcolor{brown}{SafeConfig}} (subgoal) \\
    & \texttt{\textbf{RG\_Fail}: \textcolor{brown}{FailureGoal}} (subgoal) \\
    transitions & 
    \texttt{safe(S\textsubscript{i}) \&\& safe(S\textsubscript{i+1})} (success) \\
    & \texttt{NOT safe(S\textsubscript{i}) || NOT safe(S\textsubscript{i+1})} (fail) \\
    \bottomrule
    \end{tabular}
}
\caption{Schema for \texttt{ReturnGuardMechanism} from the TMK-Structured model of the Guards and Prisoners problem taught in an AI course, specifying the finite state machine (FSM) structure, parameters, and transitions. \textcolor{blue}{Blue} refer to atomic-level operations, \textcolor{brown}{Brown}: subgoals, \textbf{Bold}: FSM state names, safe(S\textsubscript{i}): domain-specific safety condition defined over the configuration for state S\textsubscript{i}.}
\label{tab:returnguard-schema}
\end{table}

\begin{figure}[t]
  \centering
\begin{tikzpicture}[node distance=2.2cm and 3.8cm, every node/.style={font=\small}]
  \tikzstyle{state}=[draw, rounded corners, minimum width=2.8cm, minimum height=1.0cm, align=center]
  \tikzstyle{failstate}=[draw=red, fill=red!10, rounded corners, minimum width=2cm, minimum height=1.0cm, align=center]
  \tikzstyle{successstate}=[draw=black, fill=green!10, rounded corners, minimum width=2.8cm, minimum height=1.0cm, align=center]

  \definecolor{operationblue}{RGB}{0, 0, 180}
  \definecolor{subgoalbrown}{RGB}{140, 90, 40}

  \node[state] (S0) {\textbf{RG\_S0}\\\textcolor{operationblue}{EmbarkGuard}(b, g, c)};
  \node[state, below of=S0] (S1) {\textbf{RG\_S1}\\\textcolor{operationblue}{Cross}(b)};  \node[state, below of=S1] (S2) {\textbf{RG\_S2}\\\textcolor{operationblue}{DisembarkGuard}(b, g)};
  \node[successstate, below of=S2] (S3) {\textbf{RG\_S3}\\\textcolor{subgoalbrown}{SafeConfig}(newConfig)};
  \node[failstate, right of=S1, node distance=5.2cm] (Fail) {\textbf{RG\_Fail}\\\textcolor{subgoalbrown}{FailureGoal}()};

  \draw[->] (S0) -- (S1) node[midway, left] {\scriptsize safe(S0) $\wedge$ safe(S1)};
  \draw[->] (S1) -- (S2) node[midway, left] {\scriptsize safe(S1) $\wedge$ safe(S2)};
  \draw[->] (S2) -- (S3) node[midway, left] {\scriptsize safe(S2) $\wedge$ safe(S3)};

  \draw[->, dashed] (S0) to[bend left=15] node[pos=0.35, above, xshift=20pt, yshift=2pt] {\scriptsize $\neg$ safe(S0) $\vee$ $\neg$ safe(S1)} (Fail);
  \draw[->, dashed] (S1) -- (Fail) node[midway, above] {\scriptsize $\neg$ safe(S1) $\vee$ $\neg$ safe(S2)};
  \draw[->, dashed] (S2) to[bend right=15] node[pos=0.4, below, xshift=10pt, yshift=-7pt] {\scriptsize $\neg$ safe(S2) $\vee$ $\neg$ safe(S3)} (Fail);
\end{tikzpicture}

  \caption{A finite state machine representation of the \texttt{ReturnGuardMechanism} from the TMK-Structured model of the Guards and Prisoners problem, illustrating causal transitions (solid arrows) and failure branches (dashed) to represent procedural logic in TMK Mechanisms.}
  \label{fig:returnguard-fsm}
\end{figure}
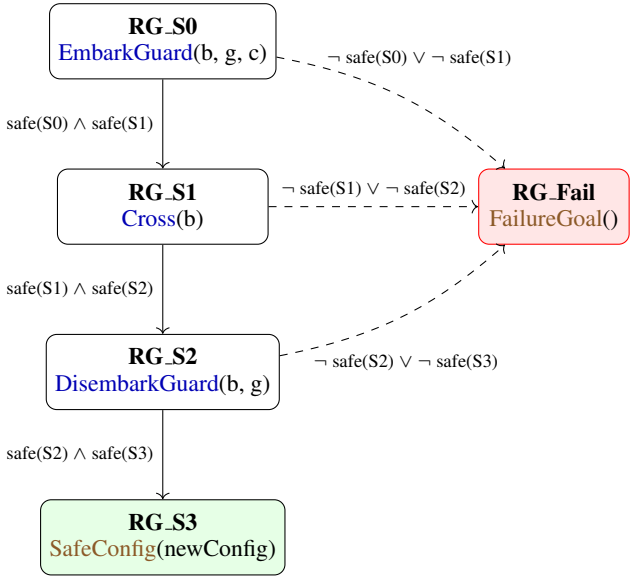

\begin{figure*}[h]
  \centering
  \includegraphics[width=\textwidth]{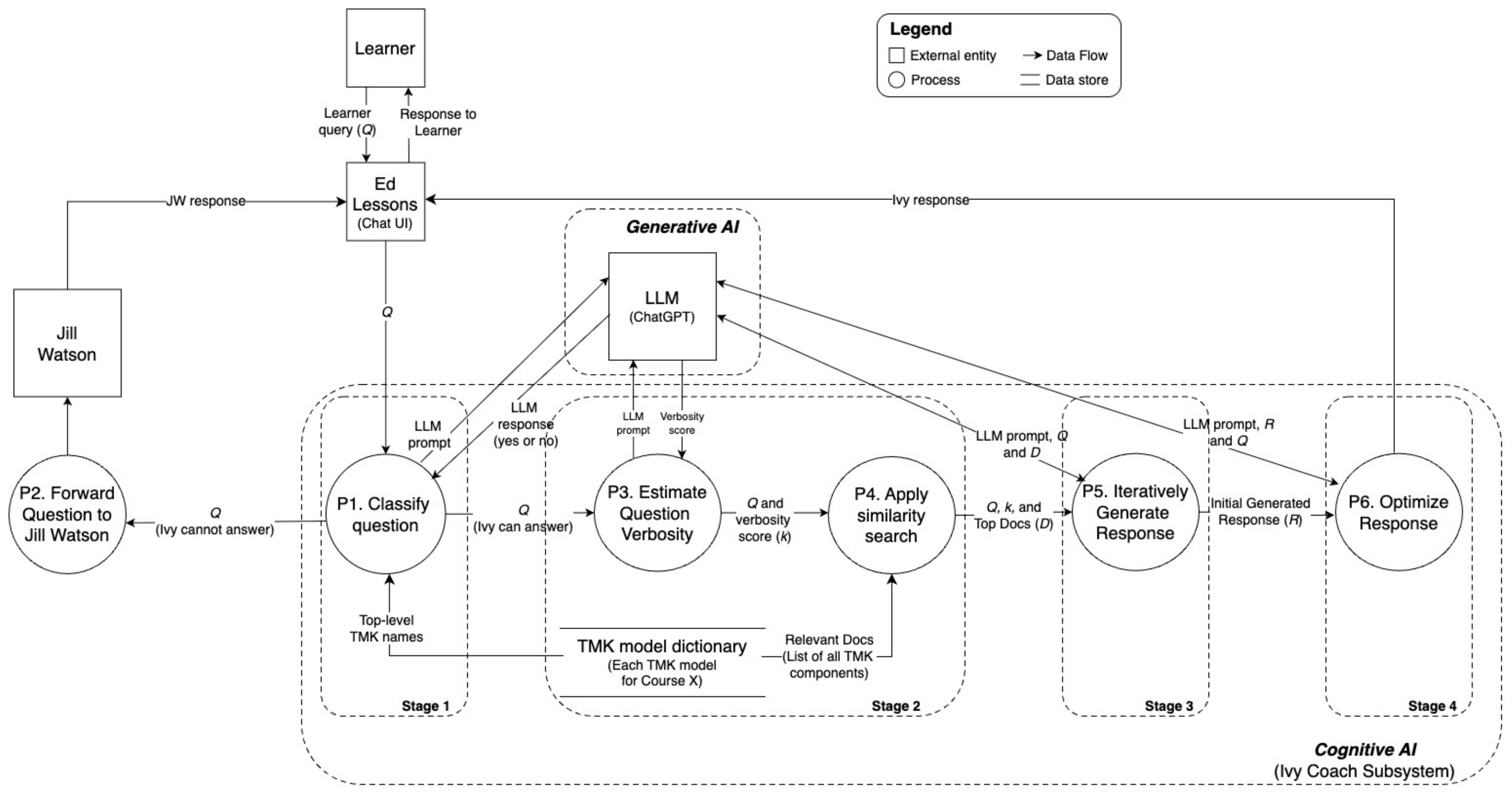}
  \caption{A four-stage constrained generation architecture for AI coaching systems. Instantiated using Ivy, an AI coach integrated within Ed Lessons to support procedural skill learning, the architecture demonstrates how an AI coach leverages symbolic control via TMK models to generate responses.}
  \label{fig:Ivy-architecture}
\end{figure*}

\section{Constrained Generation Architecture}
\subsection{Two-layered Inferencing Architecture}
We propose a two-layered inferencing architecture for AI coaching systems that support procedural skill learning. 
As an example, we ground it in a concrete skill ``Applying Semantic Networks'' to solve the classic Guards and Prisoners problem.
Suppose a learner asks: \textit{``How do we safely return a guard to the east bank without violating safety constraints?''}
Our system answers this by coordinating two kinds of inference.
A symbolic control layer encodes the permissible logic of the goal-what moves are allowed, under what conditions, and toward which subgoals-while a generative layer renders those constraints as a coherent, context-sensitive explanation (Figure \ref{fig:causal_teleo_reasoning}).

The symbolic layer provides the structure that instructors expect to see in procedural instruction: causal transitions (state-dependent guards and failure branches), teleological linkage (steps justified by the subgoals they advance), and problem decomposition (named subgoals and atomic operations).
The generative layer then verbalizes this structure so that the answer is both fluent and faithful to the underlying procedure.

This work formalizes the control layer in our proposed TMK-Structured models.
The schema (Table \ref{tab:returnguard-schema}) defines ReturnGuardMechanism as a first-class, modular component: it declares typed inputs/outputs, explicit start, success, and failure states, and a set of states that are each tagged as an operation or subgoal.
Crucially, the transitions section specifies the causal guards (e.g., \texttt{safe(S\textsubscript{i}) \&\& safe(S\textsubscript{i+1})}) that must hold for movement between states; if a guard fails, control routes to the Failure state.
Figure \ref{fig:returnguard-fsm} conceptually instantiates the schema as a finite-state machine: \texttt{RG\_S0} \(\rightarrow\) \texttt{RG\_S1} \(\rightarrow\) \texttt{RG\_S2} \(\rightarrow\) \texttt{RG\_S3(SafeConfig)} along solid, guarded arcs, with dashed arcs to \texttt{RG\_Fail} whenever a safety predicate is violated.
Teleology is explicit because completion is defined not by finishing ``some steps'', but by achieving the goal state \texttt{RG\_S3(SafeConfig)} for the parent \textit{ReturnGuard} goal. 
Decomposition is explicit because each state names the operation or subgoal it invokes (e.g., \textit{EmbarkGuard}, \textit{Cross}, \textit{DisembarkGuard}).

Contrasting this with the deployed TMK-Basic models clarifies what is novel in this paper.
In TMK-Basic, \textit{ReturnGuard} would typically appear only as a loosely described step or short paragraph: ``move a guard back while keeping the configuration safe.''
There is no named mechanism separate from the goal; no typed interface; no formal start/success/failure states; and no explicit transition guards or failure modeling.
Safety is mentioned normatively (``ensure safety''), but not operationalized as state-by-state predicates.
As a result, generative answers guided by TMK-Basic tend to be fluent yet generic: they gesture at what to do, but not why a move is legal now, how it advances the goal, or which operation is being executed.

With TMK-Structured, the generative layer is conditioned on the TMK model partially defined by Table \ref{tab:returnguard-schema} and conceptually represented in Figure \ref{fig:returnguard-fsm}.
For a ``how''-based procedural question, it is required to (i) check and narrate the relevant safety predicates before each transition (causality), (ii) connect each transition to the subgoal \textit{SafeConfig} and the parent goal \textit{ReturnGuard} (teleology), and (iii) name the invoked operation or subgoal at each state (decomposition).
A resulting explanation reads: 
\begin{quote}
``Begin in \texttt{RG\_S0(EmbarkGuard)} if \texttt{safe(S0)} holds; otherwise reconsider the passenger choice. Proceed to \texttt{RG\_S1(Cross)} while maintaining \texttt{safe(S1)}, then \texttt{RG\_S2(DisembarkGuard)} contingent on \texttt{safe(S2)}. 
This sequence achieves \textit{SafeConfig}, completing \textit{ReturnGuard} without entering the failure branch.''
\end{quote}
The wording is flexible, but the logic cannot drift: TMK model defines the valid space of moves and justifications.

Pedagogically, this separation is the point. 
By making causal guards, goal satisfaction, and decomposition explicit in the representation, TMK-Structured gives the LLM something stronger than ``relevant text'' to generate a response: it gives it the operational semantics of the skill.
The explanation that emerges is not only more specific and checkable, but also closer to the form of reasoning instructors model in class-precisely the kind of structure our evaluation later measures under the causal, teleological, and compositional dimensions.

\begin{algorithm}[t]
\caption{TMK-Guided Response Generation in Ivy}
\label{Ivy-algorithm}
\textbf{Input}: user question $Q$; TMK model $T$; language model $M$ \\
\textbf{Output}: response $R$
\begin{algorithmic}[1]
\State $TMKNames \gets$ \textsc{ExtractTopLevelNames}($T$)
\State $CanIvyAnswer \gets$ Prompt $M$ to determine if $Q$ is
\Statex \hspace*{8.3em} related to $TMKNames$
\If{$CanIvyAnswer$ = ``yes''}
    \State $\mathit{VerbosityScore}: k \gets$ Prompt $M$ to estimate
    \Statex \hspace*{10.6em} verbosity of $Q$ (range: 1--4)
    \State $TMKDocs \gets T[\mathit{Goal}] \cup T[\mathit{Mechanism}]\  \cup $
    \Statex \hspace*{8em}$T[\mathit{Concept}]$
\State $TopDocs \gets \textsc{SimilaritySearch}(Q, $
\Statex \hspace*{7em}$TMKDocs, k$)
    \State $R \gets$ empty string
    \ForAll{$d \in \mathit{TopDocs}$}
        \If{$R = ``"$}
            \State $R \gets$ Prompt $M$ to answer using $d$ and $Q$
        \Else
            \State $R \gets$ Prompt $M$ to improve answer using $d$,
            \Statex \hspace*{6.8em}$Q$ and $R$
        \EndIf
    \EndFor
    \State $R \gets$ Prompt $M$ to include only pertinent \Statex \hspace*{3.7em} information in $R$ based on $Q$
\Else
    \State $R \gets$ \textsc{RAG-GPT}($Q$)
\EndIf
\State \Return $R$
\end{algorithmic}
\end{algorithm}

\subsection{Four-Stage Pipeline and Algorithmic Implementation}
Our two-layer inferencing model (Figure~\ref{fig:causal_teleo_reasoning}) is realized as a four-stage pipeline (Figure~\ref{fig:Ivy-architecture}), formalized in Algorithm~\ref{Ivy-algorithm}.
The pipeline integrates symbolic control from TMK models with an LLM\footnote{We used GPT-4o-mini as the LLM in Ivy to ensure that our approach remains accessible and reproducible with cost-efficient models.} across four stages: (1) question scope classification, (2) structured knowledge retrieval from the TMK model, (3) constrained synthesis, and (4) coherence optimization.
We illustrate each stage using the \textit{ReturnGuard} example introduced above.

\subsubsection{Stage 1: Question Scope Classification.}
Given a learner question $Q$, we first determine whether it falls within the active skill's TMK-Structured model.
We embed the top-level names from Goal, Mechanism, and Concept components of a TMK model, and compare them to an OpenAI/FAISS embedding of $Q$.
If similarity exceeds a threshold $\tau$, the request is in-scope and proceeds to structured inference; otherwise it is routed to an unconstrained course-based RAG question-answering agent called ``RAG-GPT''.

\textit{Example.} For ``How do we safely return a guard?'', Ivy+TMK-Structured matches \textit{ReturnGuard}, \textit{ReturnGuardMechanism} and possibly other relevant documents in the TMK model, so we continue within the TMK layer.
An Ivy+TMK-Basic model would also recognize the question as in-scope, but-as we show next-it lacks an explicit formal structured representation (states, guards, success/failure), so it cannot determine or enforce which steps are permitted.

\subsubsection{Stage 2: Structured Knowledge Retrieval.}
We retrieve the top-$k$ TMK entries relevant to $Q$: the parent \emph{Goal} (inputs/requirements), the \emph{Mechanism} (finite-state machine with start/success/failure and guarded transitions), and any \emph{Concept} items (domain-specific concepts/relations such as \texttt{safe(config)}). 
Since the running example is a \textit{how}-type question, Ivy prioritizes Mechanism entries (procedural ``how-to'' knowledge) over purely declarative content; the Goal still appears in top-$k$ and is used for purpose/criteria linkage.
This yields a compact, typed specification of the operational semantics for safely returning a guard, as required by the learner's question.

\textit{Example.} Retrieval yields the \textit{ReturnGuardMechanism} schema (Table~\ref{tab:returnguard-schema})-conceptually represented by the FSM in Figure~\ref{fig:returnguard-fsm}-including safety conditions such as \texttt{safe(S\textsubscript{i}) \&\& safe(S\textsubscript{i+1})} and named atomic operations (\textit{EmbarkGuard}, \textit{Cross}, \textit{DisembarkGuard}, \textit{SafeConfig}).
The parent \textit{ReturnGuard} goal is also retrieved and provides success conditions and rationale, but the Mechanism serves as the primary scaffold for a \textit{how} answer.
In TMK-Basic, retrieval would surface descriptive text about ``sending a guard back safely,'' but not the explicit states or safe/fail transition checks.

\subsubsection{Stage 3: Constrained Synthesis.}
The LLM now answers under TMK constraints.
Rather than freely composing from snippets, it must (i) narrate only transitions specified by the Mechanism's guards (causality), (ii) connect steps to the parent goal and success state (teleology), and (iii) name the invoked operation or subgoal at each state (decomposition).

\textit{Example (abridged output).} ``Start in \texttt{RG\_S0 (EmbarkGuard)} if \texttt{safe(S0)}; otherwise revise passengers.
Proceed to \texttt{RG\_S1(Cross)} ensuring \texttt{safe(S1)}, then \texttt{RG\_S2(DisembarkGuard)} under \texttt{safe(S2)}.
This sequence achieves \textit{SafeConfig}, satisfying the \textit{ReturnGuard} goal; any violated safety predicate diverts to Failure.''
With TMK-Basic outputs are fluent but generic (e.g., ``move a guard back while maintaining safety''), because explicit state-transition constraints and requirements for stepwise justification are absent.

\subsubsection{Stage 4: Coherence Optimization.}
Finally, we compress redundant text and keep only the details needed to answer $Q$.
The optimization pass preserves TMK structure while improving readability (e.g., grouping operations under intermediate subgoals or shortening repeated guard checks).

\textit{Example.} The long, step-by-step draft is compacted to a 5–7 sentence explanation that still cites the critical guards and explicitly marks goal attainment at \textit{SafeConfig}.

We hypothesize that this four-stage design keeps the generative model expressive yet bounded.
Stages 1–2 determine whether TMK constraints apply to the learner's query and identify the relevant portions of structured knowledge; Stage 3 enforces those constraints during generation; Stage 4 polishes the output to align with instructional norms.
In the next section, we empirically discuss how the TMK-Structured model drives the measured gains in teleology and, especially, decomposition, while maintaining strong causal coherence.

\section{Structured Constraints And Inferential Quality: A System-Level Evaluation}
To assess the impact of TMK model constraints on the inferential quality of AI-generated instructional responses, we designed a structured comparative evaluation across four system\footnote{We used GPT-4o-mini for all four systems.} conditions: (1) Standard GPT (an off-the-shelf ChatGPT model), (2) RAG-GPT (GPT enhanced with retrieval-augmented generation over AI course transcripts), (3) Ivy+TMK-Basic, and (4) Ivy+TMK-Structured.
This evaluation adopts a system-level perspective focused on inferential coherence in explanations, rather than learner outcomes.

Broad benchmarks such as LogiQA \cite{ijcai2020p501} and HellaSwag \cite{DBLP:journals/corr/abs-1808-05326} evaluate general inferencing capabilities, but their breadth makes it difficult to isolate the specific cognitive competencies involved in complex problem solving \cite{ijcai2020p501,DBLP:journals/corr/abs-1808-05326,patil2025advancingreasoninglargelanguage}.
We therefore adopt a targeted, course-bound evaluation: within a graduate-level AI course, we focus on particular procedural skills.
Following principles from cognitive evaluation research \cite{evans2006heuristic}, we decompose each skill into Goals, Mechanisms, and Concepts and assess each separately \cite{kiciman2024causal}.
This design yields a granular, interpretable picture of TMK capability and supports skill-specific improvement strategies.

We curated a dataset of 20 procedural skill questions spanning four skills modeled by Ivy (five per skill).
Each system generated a response for all 20 questions, resulting in 80 total outputs.
To evaluate these responses, we developed an annotation framework centered on three dimensions of procedural inferencing:
\begin{itemize}
    \item Causality: Does the response reflect state-dependent logic or condition-based transitions between steps?
    \item Teleological: Does the response link procedural steps to broader goals or purposes?
    \item Compositional: Does the response reflect problem decomposition into operations or subgoals?
\end{itemize}

\begin{table}[h]
\centering
\renewcommand{\arraystretch}{1.1}
\begin{tabular}{c|p{6.4cm}}
\toprule
\textbf{Score} & \textbf{Definition} \\
\midrule
\textbf{0} & Absent: No evidence of the characteristic or incorrect logic. \\
\textbf{1} & Implicit / Generic: Present but vague or not grounded in the specific procedural context. \\
\textbf{2} & Explicit / Contextual: Clearly articulated and contextually relevant to the goal or model. \\
\bottomrule
\end{tabular}
\caption{Rating scale for evaluating procedural inferencing characteristics.}
\label{tab:rating-scale}
\end{table}

Ratings were provided by two types of Ivy team members: (1) a domain expert who constructed the TMK model for a given skill, and (2) independent annotators-developers familiar with TMK modeling but not involved in designing the specific model under evaluation.
Each evaluator rated generated responses using the 3-level rubric (0 = absent, 1 = implicit/generic, 2 = explicit/contextual) across the three inferencing dimensions, see Table \ref{tab:rating-scale}.
This produced a dataset of 560 system–rater evaluations (80 responses x 7 raters), each with ratings along the three inferencing dimensions.

The expert additionally annotated each question–system pair with two types of metadata: (a) a \textit{Correctness} flag (0 = incorrect, 1 = partial or unsure, 2 = correct) indicating the factual validity of the generated response, and (b) a binary (True/False) \textit{DimensionExpected} flag specifying which (one or many) of the three structural dimensions should ideally be expressed in a valid answer.

Independent annotators, in contrast, were not asked to judge correctness, but rather to assess whether structural elements were present in their observed response.
Since independent evaluations capture the perspective of novice learners without specialized familiarity of procedural skills, and expert evaluations reflect the perspective of instructors with normative expectations about inferential structures, we benchmarked three aspects: (i) which dimensions experts deemed essential, (ii) whether independents actually observed those dimensions, and (iii) whether structural ratings favored TMK-constrained systems over unconstrained baselines.

\begin{table*}[t]
\centering
\small
\setlength{\tabcolsep}{5pt}
\renewcommand{\arraystretch}{1.15}
\begin{tabular}{lcccc c cccc}
\toprule
& \multicolumn{4}{c}{\textbf{Correctness Distribution}} & & \multicolumn{4}{c}{\textbf{Inferential Dimensions (correct only)}}\\
\cmidrule(lr){2-5}\cmidrule(lr){7-10}
\textbf{System} & \textbf{Incorrect (0)} & \textbf{Partial (1)} & \textbf{Correct (2)} & \textbf{Mean$^\dagger$} & & $n_{\text{correct}}$ & \textbf{Causal} & \textbf{Teleological} & \textbf{Compositional}\\
\midrule
Ivy+TMK-Structured              & 0.20 (4) & 0.15 (3) & \textbf{0.65} (13) & 1.45 & & 13 & \textbf{1.60} & \textbf{1.333} & \textbf{1.50} \\
Ivy+TMK-Basic           & 0.10 (2) & 0.25 (5) & \textbf{0.65} (13) & 1.55 & & 13 & 1.50 & 1.167 & 1.30 \\
RAG-GPT & 0.35 (7) & 0.30 (6) & 0.35 (7)  & 1.00 & &  7 & 1.40 & 1.083 & 0.70 \\
Standard GPT   & 0.45 (9) & 0.40 (8) & 0.15 (3)  & 0.70 & &  3 & 0.70 & 0.917 & 0.60 \\
\bottomrule
\end{tabular}
\caption{Expert evaluation per system. Left: correctness distribution (proportion with count; $N{=}20$ responses per system). Right: mean ratings on three inferencing dimensions computed \emph{only over fully correct responses} (CorrectnessScore{=}2). $^\dagger$Mean correctness $= 0\cdot p_0 + 1\cdot p_1 + 2\cdot p_2$.}
\label{tab:expert-correctness-and-dims}
\end{table*}

\section{Discussion}

Our evaluation was system-level: do symbolic constraints embedded in TMK models (finite state machines, goal hierarchies, problem decompositions) make instructional explanations \emph{more correct} and \emph{more structurally reasoned}, and are those structures recognizable to novices?
Our primary hypothesis is that Ivy+TMK-Structured yields explanations with higher prevalence and quality of causal transitions, teleological linkages, and problem decomposition than Standard GPT, RAG-GPT, and Ivy+TMK-Basic.
The evaluation was intentionally not learner-outcomes-based; rather, we examined whether generated responses manifest the inferential patterns instructors intend to teach, and whether those patterns are salient to independent annotators.

\subsection{Expert-Judged Correctness}
Across 80 system outputs (5 questions × 4 skills × 4 systems; 20 per system), TMK-constrained systems substantially outperformed the baselines on expert-judged correctness.
The proportion of fully correct answers (\textit{CorrectnessScore} = 2) was 0.65 for both Ivy+TMK-Basic and Ivy+TMK-Structured, versus 0.35 for RAG-GPT and 0.15 for Standard GPT.
Error profiles reinforce this separation: TMK-Basic had the fewest outright failures (0.10 incorrect; 0.25 partial), TMK-Structured traded some partials for more zeros (0.20 incorrect; 0.15 partial), RAG was volatile (0.35 correct vs. 0.35 incorrect; 0.30 partial), and Standard GPT mostly failed or hedged (0.45 incorrect; 0.40 partial; 0.15 correct). These patterns indicate that structured constraints reduce failure rates and raise the ceiling on factual validity relative to unconstrained generation.

\subsection{Expert-Judged Inferential Structure (Means over Correct Answers)}
On the 0-2 rubric (Table \ref{tab:rating-scale}), experts rated Ivy+TMK-Structured highest on every dimension-causal 1.6, teleological 1.333, compositional 1.5-followed by Ivy+TMK-Basic (1.5, 1.167, 1.3), RAG-GPT (1.4, 1.083, 0.7), and Standard GPT (0.7, 0.917, 0.6), see Table \ref{tab:expert-correctness-and-dims}.
The largest margin appears on \emph{problem decomposition}: Ivy+TMK-Structured 1.5 vs. Ivy+TMK-Basic 1.3 ($\Delta$=+0.20), RAG-GPT 0.7 (+0.80), Standard GPT 0.6 (+0.90), aligning with TMK's design goal of scaffolding multi-step, hierarchical reasoning.
Teleology also improves under stronger constraints (Ivy+TMK-Structured 1.333 vs. Ivy+TMK-Basic 1.167; $\Delta$=+0.166), indicating clearer linking of steps to goals.
Causality is strong across TMK systems and still favors Ivy+TMK-Structured (1.6) over Ivy+TMK-Basic (1.5) and RAG-GPT (1.4), while Standard GPT lags (0.7).

\subsection{Novice validity: independents vs. expert expectations.}
Given that independents are not positioned to judge factual correctness, we do not report raw means for their ratings.
Instead, we assess validity by (i) comparing independent judgments to expert \textit{DimensionExpected} labels (affordance-aware precision/recall/F1/accuracy) and (ii) comparing independent averages to expert ratings on the same \text{QuestionId}$\times$\text{System} items.
Per-system, per-dimension agreement is highest for Ivy+TMK-Structured.
For example, teleology F1 = 0.843, Acc = 0.792; decomposition F1 = 0.752, Acc = 0.742; causality F1 = 0.765, Acc = 0.733-with Ivy+TMK-Basic close behind, RAG mid-pack (e.g., teleology F1 = 0.830, Acc = 0.775), and Standard GPT lowest on causal and decomposition (F1 $\approx$ 0.565 and 0.569).
Thus, when experts deem a structure appropriate, TMK-constrained outputs tend to express it in ways that novices can reliably recognize, evidence for both \emph{prevalence} (appears when expected) and \emph{quality} (salient, contextualized expression).

\subsection{Instructor alignment: experts' ratings vs. their expectations.}
Ivy+TMK-Structured again leads on every dimension, combining high precision and recall against the expert's own \textit{DimensionExpected} labels: causal F1 = 0.909, Acc = 0.90, Recall = 1.00; teleology F1 = 0.917, Acc = 0.90; decomposition F1 = 0.900, Acc = 0.90.
Ivy+TMK-Basic is the strongest baseline (e.g., causal F1 = 0.818; decomposition F1 = 0.800; both Acc = 0.80), with RAG and Standard GPT trailing (e.g., Standard GPT decomposition F1 = 0.526; Acc = 0.55).
Taken together with novice validity, these results show that TMK constraints make inferential structure both \emph{more present} and \emph{more interpretable}.

\subsection{Skill-level vignette (Scripts; Mechanism question: \emph{``How do scripts help AI agents understand stories?''})}
This question requires all three structures: causal transitions (how scripts \emph{produce} comprehension), teleological linkage (why each step is needed), and decomposition (how the overall process is broken down). TMK-Structured explicitly articulated the mechanism (e.g., \emph{classification} \(\rightarrow\) \emph{matching} \(\rightarrow\) \emph{gap-filling} \(\rightarrow\) \emph{interpretation}), stated the purpose of steps (teleology), and decomposed the process into analyzable components (e.g., situation classification, event matching, expectation generation). 
Baselines frequently asserted that ``scripts enable common-sense reasoning'' but lacked causal depth and decomposition.

\section{Conclusion and Future Work}

This paper introduced a constrained generation architecture that separates symbolic control from generative synthesis to improve instructional explanations for procedural skills.
Ivy operationalizes this design by using TMK models to encode causal transitions, teleological linkages, and problem decompositions, while an LLM performs context-sensitive synthesis within those constraints.
Unlike retrieval-only pipelines, TMK models act as \emph{active scaffolds} that shape both what is said and how it is structured. 

Across correctness and affordance-aware structure, Ivy+TMK-Structured produces explanations that are both more reliable (higher expert-judged correctness; fewer failures) and more pedagogically structured (higher expert means; stronger novice agreement with expert expectations), with the largest gains in problem decomposition—the core capability TMK is designed to scaffold.
This supports our primary hypothesis that TMK-constrained generation can improve the \emph{prevalence} and \emph{quality} of causal, teleological, and compositional inferencing relative to unconstrained baselines.

\subsection{Limitations}
Our study is intentionally system-level and scoped: (i) the dataset is modest (5 questions per system over four skills), which limits generalizability; (ii) correctness was judged by one domain expert per skill, so inter-rater reliability on correctness is not measured; and (iii) we evaluate explanation quality, not learning outcomes, so effects on transfer or student performance are out of scope.

\subsection{Future Work}
We will automate TMK-Structured construction using a human-in-the-loop process to cover full courses rather than a subset of skills. 
Concretely, we aim to (a) induce goal hierarchies, finite-state mechanisms, and concept ontologies from syllabi, lesson transcripts, and worked examples; (b) auto-derive \emph{DimensionExpected} masks from model structure; and (c) validate models via unit tests and counterexample-guided refinement.
Our goal is to make TMK-supported coaching broadly deployable.
In partnership with instructors, we will co-design explanation templates and dimension rubrics, then re-deploy Ivy in live courses to study learning outcomes. 
We will evaluate whether TMK-Structured-guided responses improve skill acquisition and transfer, and analyze alignment between expert expectations and novice recognition at scale.
Insights will feed back into TMK authoring and affordance modeling.

\section{Acknowledgements}
We thank Dr. Spencer Rugaber of Georgia Tech's Design Intelligence Laboratory for his invaluable guidance on TMK model development.
This research was supported by the U.S. National Science (Grants \#2247790 and \#2112532) to the National AI Institute for Adult Learning and Online Education (aialoe.org).

\bibliography{aaai2026}

\appendix

\section{A: TMK-Structured Model for Guards and Prisoners Problem}
\label{sec:A}
Before we present key aspects of the TMK-Structured framework for modeling a skill, note that the main paper focused on the Method component for the Guards and Prisoners Problem (GPP).
Here, we complete the picture by outlining the Task and Concept components used with that Method, and we include representative JSON snippets from the GPP model that Ivy uses to generate answers to learner questions.

A Task–Method–Knowledge (TMK) model is organized across three core components and represented in standard JSON.
This appendix adapts foundational material \cite{murdock2008meta,rugaber2011tmkRM,goel2017gaia} to document the schema and usage patterns we employed.
Each component plays a distinct role:
\begin{itemize}
    \item \texttt{Task:} specifies the instructional objective (what is to be achieved), including parameters, pre/postconditions, and a pointer to its procedural realization.
    \item \texttt{Method:}  specifies the procedure that achieves the objective (e.g., a finite-state machine with transitions guarded by data conditions).
    \item \texttt{Knowledge:} specifies domain entities, predicates, and relations the Task and Method depend on.
\end{itemize}

In the subsections below, we summarize each component's purpose and schema as used in TMK-Structured model for GPP and provide representative JSON.
In this appendix, TMK values (e.g., natural-language goal descriptions) appear in \textit{italics}.
Formal TMK syntax elements, such as schema keys, function names, or mechanism references, are shown in a \texttt{sans-serif} font.

\subsection{A.1: Task Component}
\texttt{Task} entries declaratively specify the objective of the skill: high-level problem aims, parameter types, semantic constraints, and the procedural means for achieving the objective via a \texttt{mechanismReference}.
Table \ref{tab:goalTable} summarizes the schema used in the GPP TMK-Structured model.
The example below shows the JSON for the top-level task \textit{TransportPassengersAcrossRiver}.
In practice, Ivy interprets a skill's TMK model in JSON format using an LLM to generate responses to learners' questions.
\newline

\begin{lstlisting}[style=jsonstyle]
"Task": [{
    "name": "TransportPassengersAcrossRiver",
    "description": "Safely move guards and prisoners across the river.",
    "inputParameters": ["initialConfiguration: configuration"],
    "outputParameters": ["finalConfiguration: configuration"],
    "given": "safe(initialConfiguration)",
    "makes": "safe(finalConfiguration)",
    "means": [{
      "mechanismReference": "GPPsolution",
      "actualArguments": ["initialConfiguration", "finalConfiguration"]
    }]
}]
\end{lstlisting}

\begin{table}[h]
\centering
\renewcommand{\arraystretch}{1.2}
\setlength{\tabcolsep}{6pt} 
\begin{tabularx}{\linewidth}{>{\ttfamily}p{0.42\linewidth} >{\raggedright\arraybackslash}X}
\toprule
\textbf{Task Schema Field} & \textbf{Description} \\
\midrule
name               & Declarative identifier of the goal or subgoal. \\
description        & Natural-language statement of the objective to be achieved. \\
inputParameters    & Typed inputs required to invoke the goal (name:type list). \\
outputParameters   & Typed outputs/effects produced when the goal succeeds. \\
given              & Logical preconditions that must hold before the goal can be invoked. \\
makes              & Logical postconditions expected to hold after the goal is achieved. \\
mechanismReference & Link to the Mechanism that procedurally realizes this goal. \\
actualArguments    & Binding from goal parameters to the mechanism’s parameters. \\
\bottomrule
\end{tabularx}
\caption{Schema for the \texttt{Task} component in the TMK-Structured model.}
\label{tab:goalTable}
\end{table}

The \textit{TransportPassengersAcrossRiver} goal declaratively captures the overarching objective of safely transporting all passengers across the river, while satisfying the domain constraints of guard-to-prisoner ratios on either side of the river and boat capacity. 
This goal takes a single input (\texttt{initialConfiguration}) and produces an output (\texttt{finalConfiguration})-both of which are typed as \texttt{configuration}.
These types are formally declared in the \texttt{Knowledge} component (see Section A.3), which defines the structure of a configuration, including the number of guards and prisoners on each bank and in the boat.

The \texttt{given} and \texttt{makes} fields specify logical preconditions and postconditions for the goal.
For this goal, we use the expression \texttt{safe(…)}
as an alias for the declarative subgoal \texttt{SafeConfiguration(…)}.
This mapping is formally defined in the \texttt{Assertions} section of the Mechanism specification, allowing a generative agent to resolve shorthand expressions and interpret safety requirements accurately (see Section A.3 for predicate definitions).

\subsection{A.2: Method Component}
Importantly, the goal is operationalized via a \texttt{mechanismReference} to the
\textit{GPPsolution} mechanism, defined in the Method component.
This mechanism encodes a complete plan as a finite-state machine (FSM), decomposing the problem into intermediate subgoals and guarded failure branches.

The current goal formulation generalizes across problem setups: it is not tied to any specific initial or final configuration.
Thus, the model can, in principle, solve GPP even from an intermediate world state (e.g., after step 5 in a failed prior plan), provided the \texttt{given} condition holds (i.e., the configuration is safe).
This makes the goal entry reusable in broader contexts, including partial re-planning or interactive explanations of incomplete solutions.
In the next section, we describe the last component of a TMK model: Knowledge.

\subsection{A.3: Knowledge Component}
The \texttt{Knowledge} component defines the domain ontology required to interpret \texttt{tasks}, \texttt{methods} and support inferencing.
This includes three core types of knowledge: \texttt{concepts}, \texttt{instances}, and \texttt{relations}.
A summary schema is provided in Table \ref{tab:conceptTable}.

\texttt{Concepts} define the types of entities in the domain.
For example, the GPP GMK-Structured model includes \texttt{guard}, \texttt{prisoner}, \texttt{boat}, and \texttt{configuration}, that can also have a hierarchical structure (e.g., a \texttt{guard} is a type of \texttt{passenger}).
The following snippet defines the \texttt{configuration} concept:
\newline

\begin{lstlisting}[style=jsonstyle]
"Knowledge": [{
    "Concepts": [{
        "name": "configuration",
        "properties": [
            { "name": "leftBank", "type": "pair" },
            { "name": "rightBank", "type": "pair" }
    ]}
]}
\end{lstlisting}

A \texttt{configuration} represents a full world state, indicating how many guards and prisoners are on each bank.

\begin{table}[h]
\centering
\renewcommand{\arraystretch}{1.15}
\setlength{\tabcolsep}{6pt}
\begin{tabularx}{\linewidth}{>{\ttfamily}p{0.42\linewidth} >{\raggedright\arraybackslash}X}
\toprule
\textbf{Concept Schema Field} & \textbf{Description} \\
\midrule
Concept        & Ontological class: a domain entity type (e.g., \textit{boat}, \textit{passenger}). \\
superConcept   & Inheritance: declares parent–child class relationships. \\
properties     & Attribute schema: named, typed fields associated with a concept. \\
Instance       & Grounded object: a specific named instance of a concept. \\
values         & Instance bindings: property–value assignments for an instance. \\
Relation       & Binary association between concepts (e.g., \textit{carries(boat, passenger)}). \\
\bottomrule
\end{tabularx}
\caption{Schema for the \texttt{Concept} component in the TMK-Structured model.}
\label{tab:conceptTable}
\end{table}

\texttt{Instances} are concrete instantiations of concepts. 
For example:
\newline
\begin{lstlisting}[style=jsonstyle]
"Knowledge": [{
    "Instances": [{
        "concept": "configuration", "name": "initialConfiguration", 
        "values": {
            "leftBank": { "guardCount": 0, "prisonerCount": 0 },
            "rightBank": { "guardCount": 3, "prisonerCount": 3 }
        }
    ]}
]}
\end{lstlisting}

This defines the initial state in GPP where all three guards and three prisoners start on the right bank. 
Such instances are used as inputs and outputs for \texttt{tasks} and \texttt{methods}.

\texttt{Relations} describe declarative associations between domain concepts.
For example:
\begin{lstlisting}[style=jsonstyle]
"Knowledge": [{
    "Relations": [{
        "name": "carries", "domain": "boat", "range": "passenger"
    }]
]}
\end{lstlisting}

This \texttt{relation} specifies that a \textit{boat} is defined to be capable of carrying a \textit{passenger} (e.g., a \textit{guard} or \textit{prisoner}).
Unlike procedural actions, these relationships do not describe interactions or state changes, but rather define domain-level constraints and permissible associations used by other TMK components (e.g., Tasks or Mechanisms) for inferencing and validation.

By capturing these three types of structured knowledge, the \texttt{Knowledge} component provides the semantic foundation needed for consistent interpretation across \texttt{Task} and \texttt{Mechanism} components.
For GenAI agents like Ivy, this enables accurate tracing of preconditions, failure triggers, and state-based reasoning, which is crucial for skill explanation and diagnostic feedback.

\section{B: Knowledge Trace of Answers Generated by Ivy for Questions about AI Skills}
\subsection{B.1 Skill: Version Spaces}
To illustrate the four-stage pipeline in action, we walk through an example on Version Spaces which is a skill taught in an AI class, showing how Ivy applies Algorithm 1 to process a learner’s question and produce a structured, domain-grounded answer.

\textbf{Question: }\textit{Under what condition should version spaces generalize the specific model?}

\subsubsection{Stage 1 — Question Scope Classification}
The learner’s question is first checked against the TMK model for the Version Space Learning lesson.
The classifier checks the question's similarity against the TMK's top-level goals, mechanisms, and concepts.

In this case, the question can be mapped directly to components such as \textit{SpecificModels} in Knowledge, \textit{GeneralizeSpecificMechanism} in Methods and \textit{GeneralizeSpecific} in Tasks.
Therefore, Ivy classifies the question as in scope and continues with the TMK-guided reasoning process rather than handing it off to an unconstrained course-based RAG question-answering agent called ``RAG-GPT''.

Here's the Ivy output from the stage:

    \texttt{Knowledge names}: \textit{Example, Model, SpecificModel, GeneralModel, Condition}
    
    \texttt{Method names}: \textit{VersionSpaceLearning, ProcessExampleMechanism, InitializeModelMechanism, GeneralizeSpecificMechanism, etc.}
    
    \texttt{Task names}: \textit{FindOutFoodAllergiesConditions, ProcessExample, InitializeModel, GeneralizeSpecific, SpecializeGeneral, etc.}
    
    \textit{Final answer: \textbf{yes}}

\subsubsection{Stage 2 — Knowledge Retrieval}
Ivy then embeds the TMK's Task, Method, and Knowledge entries, along with the question, into a FAISS vector database.
It determines the verbosity level (\textit{k}) based on the question's phrasing.
Here, \textit{k=3} for a detailed but concise answer.

From the similarity search, Ivy retrieves the top-ranked entries with most similarities to the question:

\texttt{Task}: \textit{GeneralizeSpecific} (k=0.747): $<$Definition of the goal, input/output parameters, and given/makes constraints$>$.

\texttt{Method}: \textit{GeneralizeSpecificMechanism} (k=0.739): $<$FSM states and transitions for the generalization process, including conditions to apply generalization or terminate$>$.

\texttt{Task}: \textit{SpecializeGeneral} (k=0.734): $<$Definition of the related goal for handling negative examples by refining the general model$>$.

From above we can see it retrieves two structured goals and one mechanism with FSM in Method component, as this question is more of a ``how'' question instead of simple ``what'' question.
The question contains the key action ``generalize'' and references the ``SpecificModel,'' directly matching the ``GeneralizeSpecific'' goal and mechanism.
``SpecializeGeneral'' is also retrieved because the embedding space associates generalization and specialization as complementary operations in version space.
These form the structured foundation for the response.

\subsubsection{Stage 3 — Constrained Response Generation}
Using the retrieved TMK entries as scaffolding, Ivy builds the answer within the structural and logical constraints encoded in the TMK. From the given conditions defined in \textit{GeneralizeSpecific} and \textit{GeneralizeSpecificMechanism}, the model knows that generalization occurs when input example is positive, the specific model exists, and it does not already include the example.

Ivy Output:
\textit{The version space should generalize the specific model under the following conditions:}

\textit{1. The example is a positive example (`example.isPositive == true`).}

\textit{2. The specific model is not null (`specificModel != null`).}

\textit{3. The specific model does not already include the positive example (`specificModel.includes(example) != true`).}

\textit{When these conditions are met, the generalization process can effectively expand the specific model to include the new positive example, thereby accommodating additional valid instances that fit within the learned concept. If the specific model already includes the example, no generalization is necessary, and the process can conclude successfully.}

\textit{Additionally, it is important to note that the version space also involves a specialization process based on negative examples. When a negative example is encountered (`example.isPositive != true`), the general model should be specialized to exclude this example, ensuring that the learned concept remains accurate and does not encompass invalid instances.}

\textit{In summary, the conditions for generalizing the specific model are:
}

\textit{1. The example is positive (`example.isPositive == true`).}

\textit{2. The specific model is not null (`specificModel != null`).}

\textit{3. The specific model does not already include the example (`specificModel.includes(example) != true').}

\textit{When these conditions are satisfied, the generalization process can proceed effectively, while the specialization process can be applied to negative examples to refine the general model.}

\subsubsection{Stage 4 — Coherence Optimization}
The previous response correctly identified the conditions in a structured way and provided reasoning for the necessities of the conditions. However, the explanation was verbose and contained notable redundancy.

In this stage, Ivy refines the generated answer for clarity and concision, removing repetition while keeping the essential logic. The final answer reads:

\textit{The version space should generalize the specific model under the following conditions:}

\textit{1. The example is positive (`example.isPositive == true`).}

\textit{2. The specific model is not null (`specificModel != null`).}

\textit{3. The specific model does not already include the positive example (`specificModel.includes(example) != true`).}

\textit{When these conditions are met, the specific model can be expanded to include the new positive example, allowing for a broader representation of the learned concept. If the specific model already includes the example, no generalization is needed.}

\textit{Additionally, when encountering a negative example (`example.isPositive != true`), the general model should be specialized to exclude this example, ensuring the accuracy of the learned concept.}

\textit{In summary, generalization occurs when the example is positive, the specific model exists, and it does not already include the example.}

This stage ensures the explanation is clear, logically ordered, and directly responsive to the learner’s question.

\subsection{B.2 Skill: Analogical Reasoning}

Below is another example, for the \textit{Analogical Reasoning} skill.

\textbf{Question: }\textit{What are the requirements for the 'mapping' operation?}

\subsubsection{Stage 1 — Question Scope Classification}
The first step is scope classification. A prompt-based classifier determines whether this input question aligns with the TMK model's scope by assessing relevance to Task, Method, or Knowledge components.
Based on classification results, questions are routed to either structured TMK-based reasoning for in-scope queries, or RAG-GPT for out-of-scope queries.

In this case, the question is directly concerned with the 'mapping' operation, which is explicitly defined in the TMK entries for the Analogical Problem Solving process.
At the Task level, this is represented by \textit{MappingIsValid}, which ensures structural mapping success.
At the Method level, \textit{MapStructuresMechanism} validates the mapping between the target and source structures through defined operational states.
The Knowledge level includes concepts such as structural similarity and mapping validation criteria.

Therefore, the classifier identified this question as in-scope, and its binary output (yes) indicated that it should be processed using TMK-based reasoning.
The following is Ivy output:

\texttt{Knowledge}: \textit{object, problem, solution, case, memoryStore, similarityType, superficialSimilarity, deepSimilarity, semanticSimilarity, pragmaticSimilarity, structuralSimilarity}.

\texttt{Method}: \textit{AnalogicalReasoningMechanism, RetrieveSourceCaseMechanism, MapStructuresMechanism, TransferKnowledgeMechanism, EvaluateCandidateSolutionMechanism, StoreNewCaseMechanism, SearchMemory, ValidateMapping, GenerateCandidateSolution, TestCandidateSolution, AddCaseToMemory}

\texttt{Task}: \textit{AnalogicalProblemSolving, SolveRadiationProblem, SolveAtomicStructureProblem, RetrieveSourceCase, MapStructures, TransferKnowledge, EvaluateCandidateSolution, StoreNewCase, FailureTask, SourceCaseRetrieved, MappingIsValid, CandidateSolutionGenerated, SolutionVerified, CaseStored, ProblemSolvedStored}

\textit{Final answer: \textbf{yes}}

\subsubsection{Stage 2 — Knowledge Retrieval}
Following successful scope classification, the system constructed a semantic search using the question and all TMK entries.
These were embedded and stored in a FAISS vector database. The system retrieved the most relevant entries:

\texttt{Task}: \textit{MappingIsValid} (k=0.708): $<$Definition of the goal, input/output parameters, and given/makes constraints$>$.

\texttt{Method}: \textit{MapStructuresMechanism} (k=0.700) $<$FSM states and transitions for the generalization process, including conditions to apply generalization or terminate$>$.

These components capture the operational and procedural definitions required to answer the question. 
`MappingIsValid' specifies the validation logic for confirming structural mapping success. 
`MapStructuresMechanism' defines the input/output parameters, operational states, and transitions needed to carry out the mapping.
The following is an abridged output generated by Ivy:

\textit{(Document(page\_content="{'name': 'MappingIsValid', 'description': 'Ensure structural mapping succeeded.', 'inputParameters': ['okIn: BOOLEAN'], 'outputParameters': ['okOut: BOOLEAN'], 'given': 'okIn', 'makes': 'okOut = okIn', 'means': []}", metadata={'source': 'Task', 'seq\_num': 10, 'entry': 'MappingIsValid'}), 0.7083061711688932)}

\textit{(Document(page\_content="{'name': 'MapStructuresMechanism', 'description': 'Validate structural mapping between target \& source.', 'inputParameters': ['tp: problem', 'sc: case'], 'outputParameters': ['ok: BOOLEAN'], 'organizer': {'startState': 'MS\_Check', 'successState': 'MS\_Validate', 'failureState': 'MS\_Fail', 'states': [{'name': 'MS\_Check', 'goalInvocation': {'goalReference': 'ValidateMapping', 'type': 'operation', 'actualArguments': ['tp', 'sc', 'tmpOK']}}, {'name': 'MS\_Validate', 'goalInvocation': {'goalReference': 'MappingIsValid', 'type': `goal', 'actualArguments': ['tmpOK', 'ok']}}, {'name': 'MS\_Fail', 'goalInvocation': {'goalReference': 'FailureTask', 'type': 'goal', 'actualArguments': []}}], 'transitions': [{'sourceState': 'MS\_Check', 'targetState': 'MS\_Validate', 'dataCondition': 'tmpOK'}, {'sourceState': 'MS\_Check', 'targetState': 'MS\_Fail', 'dataCondition': '!tmpOK'}]}}", metadata={'source': 'Mechanism', 'seq\_num': 2, 'entry': 'MapStructuresMechanism'}), 0.7003273091304472)}

\subsubsection{Stage 3 — Constrained Response Generation}

\textnormal{The retrieved components were used as templates for response generation. The generative language model operated within the TMK framework, ensuring the answer reflected the exact procedural and parameter requirements.}

\textit{The 'mapping' operation requires the following:}

\textbf{Requirements:}
\begin{enumerate}
    \item \textbf{Input Parameters:}
    \begin{itemize}
        \item \texttt{tp: problem} – Target structure to validate.
        \item \texttt{sc: case} – Source structure being mapped.
        \item \texttt{tmpOK: BOOLEAN} – Indicates if the initial mapping check succeeded.
    \end{itemize}
    \item \textbf{Output Parameter:}
    \begin{itemize}
        \item \texttt{ok: BOOLEAN} – True if mapping is valid, false otherwise.
    \end{itemize}
    \item \textbf{Operational States:}
    \begin{itemize}
        \item \textbf{MS\_Check}: Invokes \texttt{ValidateMapping} to check mapping.
        \item \textbf{MS\_Validate}: If \texttt{tmpOK} is true, invokes \texttt{MappingIsValid} and sets \texttt{ok}.
        \item \textbf{MS\_Fail}: If \texttt{tmpOK} is false, invokes \texttt{FailureTask}.
    \end{itemize}
    \item \textbf{Transitions:}
    \begin{itemize}
        \item \textbf{MS\_Check} $\rightarrow$ \textbf{MS\_Validate} when \texttt{tmpOK} is true.
        \item \textbf{MS\_Check} $\rightarrow$ \textbf{MS\_Fail} when \texttt{tmpOK} is false.
    \end{itemize}
\end{enumerate}

In summary, the mapping operation validates structural correspondence between a target and source, relying on specific inputs, an intermediate check, and clearly defined state transitions to ensure correctness.

\subsubsection{Stage 4 — Coherence Optimization}

\textnormal{The final step refined the response for clarity and conciseness, maintaining all procedural and structural details. Here is the final answer:}

\textit{The `mapping' operation ensures structural validation between a target and a source. It requires three inputs (\texttt{tp}, \texttt{sc}, and \texttt{tmpOK}), produces a single boolean output (\texttt{ok}), and progresses through defined operational states. These states include checking the mapping (\textbf{MS\_Check}), validating it if successful (\textbf{MS\_Validate}), or marking failure (\textbf{MS\_Fail}). Transitions between states are determined by the value of \texttt{tmpOK}. This structured approach guarantees that the mapping process is both thorough and logically consistent.}

\end{document}